\def\year{2021}\relax
\documentclass[letterpaper]{article} 
\usepackage{aaai21}  
\usepackage{times}  
\usepackage{helvet} 
\usepackage{courier}  
\usepackage[hyphens]{url}  
\usepackage{graphicx} 
\usepackage{natbib}  
\usepackage{caption} 
\usepackage{diagbox}
\usepackage{multirow, makecell}
\usepackage{amsmath}
\usepackage{multicol}
\usepackage{subcaption}
\usepackage[switch]{lineno}

\title{Directed Graph Representation through Vector Cross Product}
\author{
Ramanujam Madhavan, 
Mohit Wadhwa
\\
}
\affiliations{
	Artificial Intelligence, LinkedIn\\
	rmadhavan@linkedin.com, mwadhwa@linkedin.com
}
\begin{document}
	\maketitle
	\vspace*{-0.8cm}
	\begin{abstract}
		Graph embedding methods embed the nodes in a graph in low dimensional vector space while preserving graph topology to carry out the downstream tasks such as link prediction, node recommendation, and clustering. These tasks depend on a similarity measure such as cosine similarity and euclidean distance between a pair of embeddings that are symmetric in nature and hence don't hold good for directed graphs. Recent work on directed graphs, HOPE, APP, and NERD, proposed to preserve the direction of edges among nodes by learning two embeddings, source and target, for every node. However, these methods don't take into account the properties of directed edges explicitly. To understand the directional relation among nodes, we propose a novel approach that takes advantage of the non-commutative property of vector cross-product to learn embeddings that inherently preserve the direction of edges among nodes. We learn the node embeddings through a siamese neural network where the cross product operation is incorporated into the network architecture. Although cross product between a pair of vectors is defined in 3-dimensional, the approach is extended to learn N-dimensional embeddings while maintaining the non-commutative property. In our empirical experiments on three real-world datasets, we observed that even very low dimensional embeddings could effectively preserve the directional property while outperforming some of the state-of-the-art methods on link prediction and node recommendation tasks.
	\end{abstract}
	
	\section{Introduction}
	A growing number of real-world applications dealing with large networks such as social-networks, publication networks, are articulated as graph problems where the relationship between entities, like connection among users of a social network, are represented as nodes (e.g., members of a social-network, authors in a publication network) and edges (e.g., followee-follower relation, citation relation). These networks can range from hundreds of nodes to millions or billions of nodes, and understanding the complex relationships among large graphs is a challenging problem in the industry as well as academia. To understand and capture the relation between nodes in large graphs, graph embedding techniques are commonly used that map each node in a graph to a low dimensional vector space. Machine learning methods that are well-defined and applied on vector space can then be leveraged on these node embeddings to do tasks like link prediction \cite{liben2007link}, recommendation \cite{ying2018graph}, clustering, etc. A classic example is to use cosine similarity between a pair of node embeddings to predict if a link exists between them. 
	
	To embed the node in low dimension vector space, various graph embedding methods have been proposed in the literature. The well-known methods, like DeepWalk ~\cite{perozzi2014deepwalk}, Node2Vec ~\cite{grover2016node2vec}, consider random walks to understand the neighborhood of the node and use a skip-gram model training mechanism to learn node embeddings. These methods perform well for undirected graphs and can be scaled to large graphs. Another well-known method, LINE ~\cite{tang2015line}, considers first and second order proximity of the node with the optimization task to minimize the KL divergence between input graph information and the output embeddings. However, these methods target undirected graphs, and directional relationships among entities are not well attended.
	
	Directional relation occurs naturally in various real world applications like twitter network, citation network, and is studied by the following well-known methods, HOPE ~\cite{ou2016asymmetric}, APP~\cite{zhou2017scalable} and NERD ~\cite{khosla2019node}. HOPE preserves asymmetric transitivity relations by approximating high-order proximity, whereas APP is based on random walks strategy. Both the methods learn two embeddings for each node, namely source and target embedding, and the direction of edge among a node pair (say node X to node Y) is predicted based on source embedding of node X and target embedding of node Y. These methods try to learn the neighborhood and direction property of the network with source and target embeddings for each node. Therefore, it tries to balance between the two optimization tasks, distance and direction, that may impact performance.
	
	A valuable property to ensure for a directed edge, say edge exists for node X to node Y but not the other way, is that if the method predicts edge for node X to node Y then it should strictly predict no edge for node Y to node X, that is, we are looking for a non-commutative operator that can be applied on the node embeddings to predict edge for one way and strictly no for other way in case of a directed edge.
	
	To address the above shortcomings of HOPE, APP and NERD and ensuring the non-commutative property in case of directed edge, we propose a novel method, \emph{Graph Representation Encoding Edge Direction} (GREED), that builds on existing work like DeepWalk ~\cite{perozzi2014deepwalk}, Node2Vec~\cite{grover2016node2vec} to capture the neighborhood property of the graph and learns a respective embedding for each node to capture the direction property. GREED learns embedding with the cross-product operator that ensures that the non-commutative property is preserved for directed edges and for bi-directional edges, the operator assigns near equal prediction probability to edges in both directions.
	
	GREED outperforms other methods in terms of understanding the directional relation between the nodes and is able to learn the direction embeddings in a very low dimensional vector space. We evaluate our approach on link prediction and node recommendation tasks on three real-world datasets and compare the performance with other well known methods, also state-of-the-art methods in some case, such as DeepWalk, Node2Vec, LINE, HOPE, NERD. The results show that our approach outperforms the existing methods by a significant margin.
	
	\section{Related Work}
	Graph embeddings are studied in detail for undirected graphs. Some of the early approaches applied dimensionality reduction techniques on the adjacency matrix of graphs. Holland and Leinhardt~\shortcite {holland1981exponential} described an exponential family of probability distribution to capture the structural properties in directed graphs, including the attractiveness and expansiveness of nodes and the reciprocation of edges. As an extension to the above model, Wang and Wong~\shortcite{wang1987stochastic} proposed Blockmodels to capture subgroup information of nodes. 
	
	More recent work on graph modeling are towards scalable approaches. DeepWalk ~\cite{perozzi2014deepwalk} presented a random walk based strategy to capture the relation among vertices, which is similar to the relationship between words in a sentence. The random walks are then passed to word2vec skip-gram model to learn the representation of the vertices. Node2Vec ~\cite{grover2016node2vec} improvised on random walk procedure to efficiently explore and capture the node neighborhoods using breadth-first and depth-first search traversals. LINE~\cite{tang2015line} is another approach that builds a representation to preserve both local and global network structure. While these approaches are scalable, they don’t preserve the directionality of edges.   
	
	When it comes to modeling directed graphs, Perrault-Joncas and Meila ~\shortcite{perrault2011directed} and Mousazadeh and Cohen~\shortcite{mousazadeh2015embedding} explored how diffusion kernel determines the overall connectivity and asymmetry of the resulting graph and demonstrated how Laplacian-type operators can offer insights into the underlying generative process. Chen and Yang \shortcite{chen2007directed}  learn the embedding vectors in Euclidean space using the transition probability together with the stationary distribution of Markov random walks to measure locality property.  However, all of these methods do not scale to large graphs.
	
	HOPE~\cite{ou2016asymmetric} preserved asymmetric transitivity by proposing a high order proximity preserved embedding method by deriving a general formulation covering multiple high order proximity measures such as Katz measure and Rooted PageRank. The method also provided an upper bound for the approximation error. However, tying to a particular measure comes at the cost of generalization. APP~\cite{zhou2017scalable} proposed a random walk based method to encode rooted PageRank proximity. It uses directed random walks with restarts to generate training pairs and trains a pair of source and target embedding for every node to capture directionality. NERD~\cite{khosla2019node} extends the random walks to cases where nodes have zero outdegree and general random walks startegies get stuck there. The extension is based on alternating walks between source and target nodes and therefore it learns two embeddings for each node.
	
	\section{Graph Representation Encoding Edge Direction Approach}
	\subsection{Motivation}
	A desirable embedding model should satisfy proximity relations among entities along with other useful attributes such as edge direction between node pairs in a graph, antonym/synonym relationship between co-occurring words, hierarchy between entities. For example, if a word embedding is able to provide antonym or synonym relation between a pair of words in addition to co-occurrence relation or an entity embedding is able to provide hierarchy among entities in addition to similarity, they would be immensely useful in certain use cases. While the distance metric between a pair embeddings represents the similarity, their direction of alignment with respect to each other can be exploited to encode such additional attributes. Therefore, the motivation here is to expand the concept of embedding from capturing mere semantic similarity by proximity into capturing additional useful attributes about the relationship between entities. In the realm of directed graphs, our objective is to encode the direction of the edge between nodes into their embeddings by directionally aligning them in a certain way. 
	
	While learning the graph node embeddings, typically a distance measure between embeddings is optimized to predict the appropriate label. To enforce the alignment discipline into embeddings, we need the necessary additional components and the mathematical constructs in the learning framework. We introduce \emph{Graph Representation Encoding Edge Direction} (GREED), that incorporates these additional constructs to solve this problem. The following section discusses the method in detail.
	
	\begin{figure*}[tb]
		\centering
		\includegraphics[scale = 0.6]{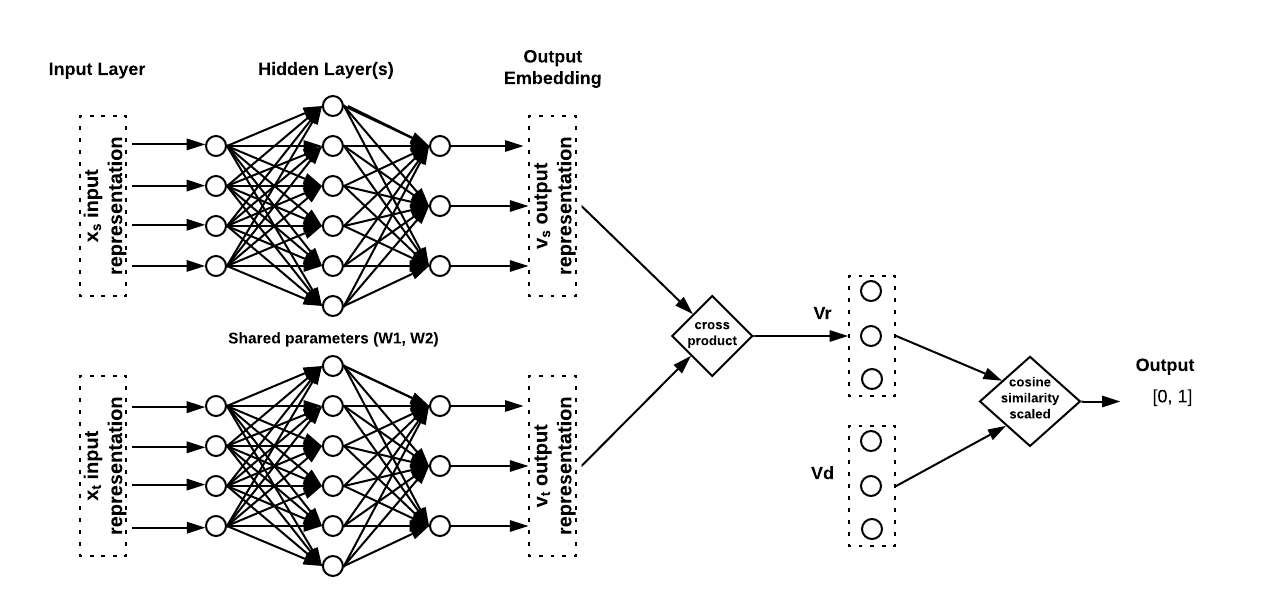}
		\caption{GREED Network Architecture}
		\label{fig:GNA}
	\end{figure*}
	
	\subsection{Problem Formulation}
	\subsubsection{Proximity Representation} The first objective of a graph embedding model is to encode the proximity of nodes. Random walk based methods along with language modeling techniques have achieved state-of-the-art performance on numerous real-world undirected graph datasets. DeepWalk, Node2Vec, and other methods that are based on this technique have proved to be scalable on large graphs. Therefore, we leverage the embeddings from the existing methods for proximity understanding.
	
	\subsubsection{Direction Representation}
	Given the understanding of proximity, our next objective is to learn a direction embedding in such a way that an operator, when applied on a pair of nodes within proximity, will help infer the edge direction. In other words, given a pair of direction embeddings $v_s$ and $v_t$ for nodes $s$ and $t$ respectively, we need an indicator function, $\delta(v_s, v_t)$, that satisfies the non-commutative property, $\delta(v_s, v_t) \neq \delta(v_t, v_s)$, for a directed edge between the node pair. Since the cross product operator is non-commutative in vector space, we formulate the indicator function as a function of cross product between $v_s$ and $v_t$ as:
	\begin{equation}
	\delta (v_s,v_t) = \varphi (v_s \times v_t)  
	\end{equation}
	
	\noindent  In addition being non-commutative, a key property of cross product is that its output is a vector that is orthogonal to the plane containing the operand vectors and we optimize the direction of this vector to infer the desired property about the edge between the operand vectors. As the direction in space can only be inferred relative to a reference, we introduce an arbitrary but fixed reference vector $v_d$ against which the alignment will be measured and optimized during training. With this idea, we expand and modify equation 1, into equation 2, to be a function of the resulting cross product output, $v_r = v_s \times v_t$, and the reference vector $v_d$ that calculates the cosine similarity measure $\phi$ (equation 4) followed by similarity value scaling  to produce output between $[0,1] $ (equation 3) and applies a threshold $\tau$ to produce a binary result. 
	
	\begin{equation}
	\delta_{v_d} (v_s, v_t)=\left\{
	\begin{array}{ll}
	1 & \quad   \varphi_{v_d}  ( v_s \times v_t  ) > \tau \text{ ,}   \\
	0 & \quad {else}
	\end{array}
	\right.
	\end{equation}
	
	\begin{equation}
	\text{where  \space \space} \varphi_{v_d}  ( v_r )   = \frac{ 1 +   \phi  (v_d, v_r )}{2} \text{ ,}
	\end{equation}
	
	\begin{equation}
	\text{\space \space \space \space \space \space \space \space \space} \phi  (v_d, v_r )   =   \frac {v_d \cdot v_r}{||v_d|| ||v_r||}   
	\end{equation}

	\noindent Note that $v_r$ here is defined for 3-dimensional space and we generalize the approach for N-dimensional space in the section below. 
	
\begin{figure}[h!]
		\begin{subfigure}[b]{0.495\linewidth}
			\captionsetup{justification=centering}
			\includegraphics[scale=0.20]{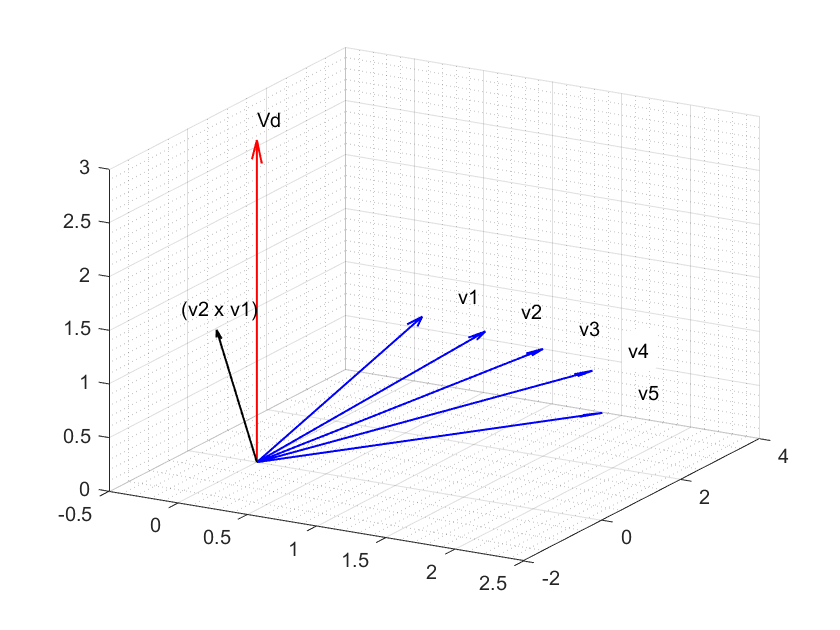}
			\caption{Cross Product $v_2 \times v_1 $}
		\end{subfigure}
		\begin{subfigure}[b]{0.495\linewidth}
			\captionsetup{justification=centering}
			\includegraphics[scale=0.20]{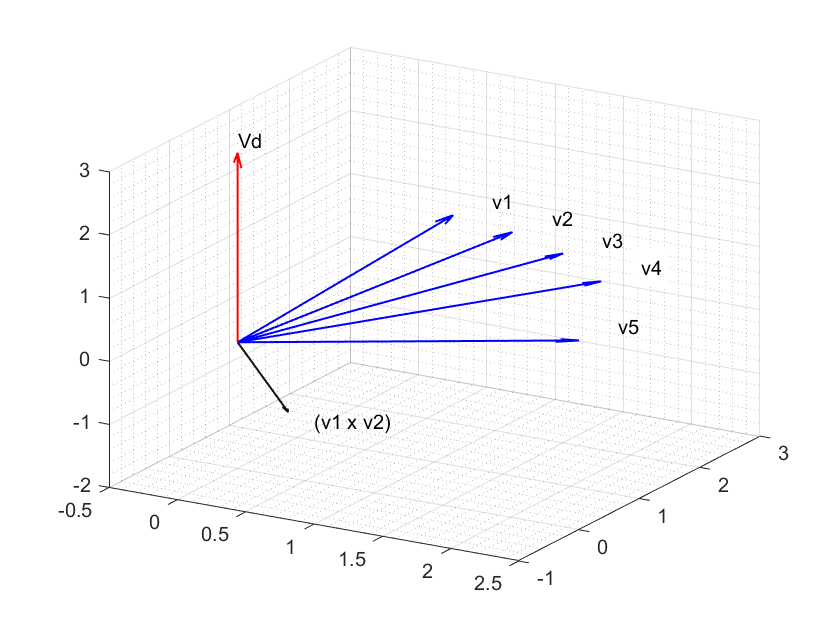}
			\caption{Cross Product $v_1 \times v_2 $}
		\end{subfigure}\par
		\caption{Illustration of sample embeddings ($v_1$, $v_2$, $v_3$, $v_4$, $v_5$) with respect to reference vector $v_d$ in GREED model}
		\label{fig:CPE}
	\end{figure}

	Figure \ref{fig:CPE} provides a visual intuition for the spatial arrangement of direction embeddings for a set of sample vectors in 3D space. In figure \ref{fig:CPE}a, the cross product of $v_2$ with $v_1$ makes an acute angle with $v_d$ while the cross product of $v_1$ with $v_2$ makes an obtuse angle with $v_d$ in figure \ref{fig:CPE}b.  By defining a threshold on this angle, we infer a specific attribute of the relationship between $v_1$ and $v_2$. 
	\subsection{GREED Model}
	With the above problem formulation, we now present the GREED model that is built on siamese network \cite{bromley1994signature} followed by cross-product and cosine similarity operations optimized for learning direction embeddings with contrastive loss. For proximity understanding, we leverage one of the existing graph embedding methods for (un)directed graphs. In a two step process, we first infer the presence of edge between a pair of nodes from the proximity embeddings and then infer the direction from the direction embeddings. We augment the given graph edges with random walks to further learn transitive edge directions. The random walks are limited to close proximity with respect to node, similar to DeepWalk, LINE.
	
	\subsubsection{Siamese Network} 
	We use siamese network to learn a direction embedding for every node by optimizing for alignment between pair of node embeddings and taking cross-product anti-commutative property into account that yields the constraint in equation 5 on the prediction value.
	The constraint allows the model to learn prediction value close to 1 for directed edge from node $s$ to node $t$ and naturally keep prediction value close to 0 for the reverse no-edge case. For the undirected edge, the model learns to assign near equal (close to $0.5$) prediction value for both the directions.
	\begin{equation}
	\varphi_{v_d}  ( v_s \times v_t ) = 1 - \varphi_{v_d} (v_t \times v_s)
	\end{equation}
	
	\noindent Figure \ref{fig:GNA} shows the network structure where we limit the network to one hidden layer for the scope of explaining the method but in practice the network could be extended to more hidden layers. The inputs are the $K$-dimensional initial feature vectors (one-hot encoding, node attribute vector or randomly initialized vector) of the source and target nodes, $x_s,\: x_t \in R^K$ respectively. The output layer of the fully connected siamese network provides the low $N$-dimensional node embeddings for source and target  nodes, $v_s,\: v_t \in R^N$ respectively. $W_1 \in R^{K \times L}$ and $W_2 \in R^{L \times N}$ are the learnt parameters associated with the hidden layer (of size $L$) and embedding output layer respectively. We introduced the cross product operation in the network between $v_s$ and $v_t$ to produce the vector $v_r \in R^N$. The cosine similarity between $v_r$ and $v_d$ is scaled to produce the output prediction. We backpropagate the contrastive loss of this output against the true label and optimize the network parameters through gradient descent. We use $ReLU$ as the activation function on the siamese hidden layer(s). Activation is not applied on the last layer that produces the embedding. Although, we set $v_d$ to an arbitrary fixed vector, it can also be learnt in the backpropagation process.
	
	Forward-propagation pass of the above network is presented below in the set of equations 6.
	\begin{equation*}
	a_s = ReLU(W_1x_s) \quad \text{and} \quad a_t = ReLU(W_1x_t)
	\end{equation*}
	\begin{equation*}
	v_s = W_2a_s \quad \text{and} \quad v_t = W_2a_t
	\end{equation*}
	\begin{equation*}
	v_r = v_s \times v_t
	\end{equation*}
	\begin{equation}
	\hat y = \frac{1 + \frac {v_d \cdot v_r}{||v_d|| ||v_r||}}{2}
	\end{equation}
	
	\subsubsection{Contrastive Loss}
	We apply contrastive loss~\cite{hadsell2006dimensionality}  to the output defined in equation 6 against the true edge label $y$ and backpropagate the loss through the network. Contrastive loss is chosen to provide greater degree of freedom in terms of alignment to pairs of node embeddings with undirected edges by choosing margin $\rho$ appropriately. 
	\begin{equation}
	L(\hat y, y)   = y ( 1-\hat y) ^2 + (1-y) . max ( \hat y - \rho, 0) ^2
	\end{equation}
	Given equation 5 and 7, margin $\rho$ should be kept $< 0.5$ for GREED network as it allows to optimize for undirected edge prediction value between $(\rho, 1 - \rho)$ and directed edge prediction value at least $1 - \rho$.
	
	\subsubsection{Back Propagation}
	The network we present with cross-product operation is novel to the best of our knowledge and hence for completeness, we elaborate in this section the gradient computations required to backpropagate the contrastive loss to learn the parameters $ W_1 $ and $W_2$. Equations 8 and 9 provide the partial derivate of the loss with respect to $ W_2 $ and $W_1$ respectively.
	\begin{equation}
	\frac{\partial L}{\partial W_2}  = \frac{\partial L}{\partial \hat y} \cdot \frac{\partial \hat y}{\partial v_r} \cdot \frac{\partial v_r}{\partial W_2}
	\end{equation}
	\begin{equation}
	\frac{\partial L}{\partial W_1}  = \frac{\partial L}{\partial \hat y} \cdot \frac{\partial \hat y}{\partial v_r} \cdot \frac{\partial v_r}{\partial a_s}  \cdot \frac{\partial a_s}{\partial W_1} + \frac{\partial L}{\partial \hat y} \cdot \frac{\partial \hat y}{\partial v_r} \cdot \frac{\partial v_r}{\partial a_t}  \cdot \frac{\partial a_t}{\partial W_1}
	\end{equation}
	\newline
	Where $a_s$ and  $a_t$ are the ReLU activated output of the hidden layer of the siamese network and $W_2a_s$ and  $W_2 a_t$ are the output embeddings $ v_s $ and $ v_t$ respectively. We express the components of equation 8 and 9 below from equation 10 till equation 17.
	\newline
	\begin{equation}
	\frac{\partial L}{\partial \hat y}  = \left\{
	\begin{array}{ll}
	- 2 (1 - \hat y) & \quad y=1 \\
	2(\hat y - \rho) & \quad  y=0 \ and \ \hat y > \rho  \\
	0 & \quad y=0 \ and \ \hat y \leq \rho\\
	\end{array}
	\right.
	\end{equation}
	\begin{equation}
	\frac{\partial \hat y}{\partial v_r}  = \frac{1}{2} \left ( \frac {\partial \phi}{\partial v_r} \right )
	\end{equation}
	\begin{equation}
	\frac {\partial \phi}{\partial v_r}  = \frac {v_d }{||v_d|| ||v_r||} - \frac {v_r \cdot \phi  ( v_r, v_d )  }{||v_r||^2}
	\end{equation}
	
	\noindent We now handle the partial derivative of $v_r$ with respect to $W_2 $ and $a_s$ where $v_r = v_s \times v_t$. Note that the partial derivative of cross product adheres to the chain rule. 
	\begin{equation} \label{eq12}. 
	\begin{split}
	\frac{\partial v_r}{\partial W_2} & = \frac{\partial v_s}{\partial W_2} \times v_t + v_s \times  \frac{\partial v_t}{\partial W_2}  \\
	& =\frac{\partial (W_2 \cdot a_s)}{\partial W_2} \times v_t + v_s \times  \frac{\partial (W_2 \cdot a_t)}{\partial W_2} \\
	& = a_s \times v_t + v_s \times a_t
	\end{split}
	\end{equation}
	\begin{equation} \label{eq13}
	\begin{split}
	\frac{\partial v_r}{\partial a_s} & = \frac{\partial (v_s \times v_t)}{\partial a_s} \\
	& =\frac{\partial v_s}{\partial a_s} \times v_t + v_s \times  \frac{\partial  v_t}{\partial a_s} \\
	& =\frac{\partial (a_s \cdot W_2)}{\partial a_s} \times v_t + v_s \times  \frac{\partial  (a_t \cdot W_2)}{\partial a_s} \\
	& = W_2 \times v_t
	\end{split}
	\end{equation}
	\begin{equation} \label{eq14}
	\begin{split}
	\frac{\partial v_r}{\partial a_t} & = \frac{\partial (v_s \times v_t)}{\partial a_t} \\
	& =\frac{\partial v_s}{\partial a_t} \times v_t + v_s \times  \frac{\partial  v_t}{\partial a_t} \\
	& =\frac{\partial (a_s \cdot W_2)}{\partial a_t} \times v_t + v_s \times  \frac{\partial  (a_t \cdot W_2)}{\partial a_t} \\
	& = v_s \times W_2
	\end{split}
	\end{equation}
	\newline
	\begin{equation} \label{eq15}
	\begin{split}
	\frac{\partial a_s}{\partial W_1} & = \frac{\partial \ max( W1 \cdot x_s,0) }{\partial W_1} \\
	& =\left\{
	\begin{array}{ll}
	0 & \quad  W_1 \cdot x_s <0  \\
	x_s & \quad W_1 \cdot x_s >0 
	\end{array}
	\right.
	\end{split}
	\end{equation}
	\newline
	\begin{equation} \label{eq15}
	\begin{split}
	\frac{\partial a_t}{\partial W_1} & = \frac{\partial \ max( W1 \cdot x_t,0) }{\partial W_1} \\
	& =\left\{
	\begin{array}{ll}
	0 & \quad  W_1 \cdot x_t <0  \\
	x_t & \quad W_1 \cdot x_t >0 
	\end{array}
	\right.
	\end{split}
	\end{equation}
	
	\hfill\\
	\subsubsection{Generalized Vector Cross Product} \hfill\\
	Though cross product between two vectors is defined in 3-dimensional space, some of its familiar properties can be extended to N-dimensional space. In particular, the property we need to extend GREED to high dimensions is the non-commutative property.
	
	\noindent \textbf{Formulation:} Similar to cross product in $R^3$, the generalized cross product in N-dimensions is dependent on $N-1$ vectors that lie in $R^N$ i.e. $v_1 \times v_2 \times … \times v_{N-1}  = v_N$ where $v_1, v_2, ... , v_{N-1} \in R^N$. As with the 3-dimensional cross product, N-dimensional cross product can be computed by usual determinant formula. Non-commutative property of N-dimensional cross product $v_1 \times v_2 \times v_3 … \times v_{N-1} = - (v_2 \times v_1 \times v_3 … \times v_{N-1})$ can be trivially proved by determinant property that swapping two rows of a determinant matrix toggles the sign of the resulting value. Given the above forumulation, we can extend GREED to N-dimensional embeddings by adding $N-3$ arbitrary but fixed (mutually linearly independent) vectors along with 2 vectors (source and target) to compute the resulting cross-product in $N$-dimension. Back-propagation for N-dimensional cross product remain the same as in 3-dimension, where the arbitrary $N-3$ vectors are constants while computing the derivative (in equation 13). To limit the scope of the paper, we don't work with this formulation in the experiments section.
	
	\section{Experiments}
	In this section, we evaluate our approach on three publicly available real-world datasets. We first present the datasets information, then show the comparison of our approach with well-known state-of-the-art graph embedding methods on link-prediction and node-recommendation tasks.
	
	\begin{table}
		\centering
		\begin{tabular}{| c | c | c |}
			\hline
			Dataset & Number of Nodes & Number of Edges \\ \hline \hline
			Cora & 23,166 & 91,500 \\ \hline
			Epinion & 75,879 & 508,837 \\ \hline
			Twitter & 465,017 & 834,797 \\ \hline
		\end{tabular}
		\caption{Statistics of Datasets}
		\label{tab:datasetstats}
	\end{table}
	
	\subsection{Datasets}
	We test our approach on the following three directed graph datasets. Table \ref{tab:datasetstats} summarize some of the statistics of these datasets.
	
	\begin{itemize}
		\item \textbf{Cora} \footnote[1]{http://konect.cc/networks/subelj\_cora/} : A citation network of scientific papers where the directed edge represents a citation relation between scientific papers where each paper is a node in the network. 
		
		\item \textbf{Epinion} \footnote[2]{http://konect.cc/networks/soc-Epinions1/} : A who-trust-whom online social network of a general consumer review website. The nodes represent the users and directed edges represent the trust relation.
		
		\item \textbf{Twitter} \footnote[3]{http://konect.cc/networks/munmun\_twitter\_social/} : A directed network containing information about who follows whom on Twitter. Nodes represent users and an edge shows the follower-followee relation.
		
	\end{itemize}
	
	\begin{table*}
		\centering
		\begin{tabular}{|*{10}{c|}}
			\hline
			\multirowcell{3}{\backslashbox{Method}{Dataset}} & \multicolumn{3}{c|}{Cora} & \multicolumn{3}{c|}{Epinion} & \multicolumn{3}{c|}{Twitter} \\
			\cline{2-10}
			& \makecell{\\ Type 1 \\} & \makecell{\\ Type 2 \\} & \makecell{\\Type 3\\} & \makecell{\\Type 1\\} & \makecell{\\Type 2\\} & \makecell{\\Type 3\\}  & \makecell{\\Type 1\\} & \makecell{\\Type 2\\} & \makecell{\\Type 3\\} \\ \hline \hline
			DeepWalk & 0.7541 & 0.5183 & \textbf{0.9775} & 0.7960 & 0.6296 & 0.8908 & 0.6394 & 0.5009 & 0.7769 \\ \hline
			Node2Vec & 0.7342 & 0.5165 & 0.9403 & 0.7167 & 0.6336 & 0.7640 & 0.5016 & 0.5000 & 0.5010 \\ \hline
			LINE & 0.6490 & 0.5061 & 0.7825 & 0.5750 & 0.5433 & 0.5931 & 0.5091 & 0.5016 & 0.5150 \\ \hline
			HOPE & 0.8536 & 0.7950 & 0.9091 & \textbf{0.8265} & 0.6843 & \textbf{0.9075} & 0.7694 & 0.7689 & 0.7699 \\ \hline
			NERD & 0.8097 & 0.7563 & 0.8603 & 0.7228 & 0.5631 & 0.8138 & 0.8408 & 0.8420 & 0.8397 \\ \hline
			GREED & \textbf{0.9130} & \textbf{0.8718} & 0.9520 & 0.7949 & \textbf{0.7443} & 0.8220 & \textbf{0.9260} & \textbf{0.9962} & \textbf{0.8562} \\ \hline
		\end{tabular}
		\caption{ROC-AUC scores for Link Prediction}
		\label{tab:lpresults}
	\end{table*}
	
	\subsection{Baseline Methods}
	
	\begin{itemize}
		\item \textbf{DeepWalk} does random walks to understand the connection between nodes and the random walks are treated as sentences for word2vec skipgram model that learns to predict the adjacent nodes for a given node. 
		
		\item \textbf{Node2Vec} instead of doing random walks, does biased random walks where the walk is controlled by two parameters to prioritize breadth first search for understanding local neighborhood and depth first search for understanding global neighborhood. The training mechanism is the same as DeepWalk.
		
		\item \textbf{LINE} uses KL divergence to minimize the probability distribution over input graph adjacency matrix and output dot product of node embeddings. It takes into account first order and second order proximity where they show that second order proximity works better than first order.
		
		\item \textbf{HOPE} proposed a high order proximity preserved embedding method by deriving a general formulation covering multiple proximity measures such as Katz measure and Rooted PageRank. The implementation that is used as a baseline here uses Katz proximity measure.
		
		\item \textbf{NERD} tries to preserve directed edge property by learning two embeddings for each node same as APP. Instead of doing directed random walks, NERD does alternating random walks with the idea that the directed random walks could get stuck on nodes with zero out-degree whereas the alternating random walks covers both source and target nodes in an alternative way. The idea here is inspired from SALSA \cite{lempel2001salsa} and HITS \cite{kleinberg1999authoritative}.
	\end{itemize}
	
	\begin{table}
		\centering
		\begin{tabular}{| c | p{5cm} |}
			\hline
			Dataset Type & Edges Type \\ \hline \hline
			Type 1 & Positive Edges + Negative Edges + Random Negative Edges \\ \hline
			Type 2 & Positive Edges + Negative Edges  \\ \hline
			Type 3 & Positive Edges + Random Negative Edges  \\ \hline
		\end{tabular}
		\caption{Test Dataset Types}
		\label{tab:testdatasets}
	\end{table}
	
	\subsection{Experimental Setting}
	In our experimental setting for GREED, we leverage DeepWalk $128$-dimension embeddings on undirected graph for proximity understanding. The siamese network architecture is based on two hidden layers of $256$ dimensions with ReLU activation layer followed by a dense output layer of dimension $3$ with $\rho = 0.25$ for contrastive loss. Random walks upto $3rd$ degree neighbor is done to construct training data with positive and negative (reverse) edges. Inference prediction task on node pair embeddings for GREED is a two step process where first optimal threshold is picked for proximity embeddings based on ROC-AUC curve and then direction embedding model we proposed is applied to the node pair satisfying the proximity constraint. On recommendation task, proximity embeddings are first considered to arrange recommended nodes in terms of similarity score, and then direction prediction is done to pick top-K recommended nodes.
	
	For other methods, we set dimensions to $128$ for fair comparison with similarity measured as scaled cosine similarity (equation 3). Other settings are based on the parameter values suggested by the author or default values present in the author's implementation. Negative samples $= 5$ for LINE (with order 2) and NERD with $100M$ samples; number of walks $= 40$ with walk length $= 40$ for DeepWalk; number of walks $= 40$ with walk length $= 80$ for Node2Vec; default settings for HOPE with tolerance = $0.0001$.
	
	\subsection{Link Prediction}
	A fundamental problem in networks is to predict the possibility of link between a given node pair, and link prediction task aims to evaluate this. We conduct this evaluation by removing $20\%$ of edges randomly picked from the dataset and keeping it under the test dataset, the remaining dataset is used as a training set. Since we are interested in the performance of methods in directed graph setting, we augment the test set with two types of edges i.e. negative edges and random negative edges. Negative edges are the complement of positive edges in the test set, only if they don’t occur as positive edges in the complete network. Random negative edges are randomly sampled (equal proportion as positive edges) node pairs from the network and are denoted to have no edge between them. We create three types of test datasets to understand the effect of methods in directed graphs as presented in Table \ref{tab:testdatasets}.
	
	Receiver Operating Characteristic - Area Under Curve (ROC-AUC) scores for all methods on link prediction task are summarized in Table \ref{tab:lpresults}. We can see that, as expected existing methods perform well on Type 3 dataset which doesn't take into account the negative edges. This is because the methods based on symmetric distance measures are able to understand the proximity well. However, as expected the symmetric distance doesn't hold good for negative edges in the test dataset and we can see that the results are affected for these methods on Type 1 and Type 2 datasets. NERD which works on directed graph shows better performance on Type 1 and Type 2 datasets. Our method, GREED, is able understand the directional property well and thus outperforms other methods on Type 1 and Type 2 datasets. On twitter dataset especially, which is a large directed follower-followee network, our method is able to understand the orientation of edges in the graph and predict almost all edges, within proximity, correctly. 
	
	\subsection{Node Recommendation}
	A well-known problem area in networks is to personalize the recommendation for a given node i.e. to recommend top-k intimate nodes from a large set of nodes. This task setting is also discussed in NERD and we evaluate all methods on this task. As before, we remove $20\%$ of edges randomly picked from the dataset and keep it under the test dataset. To do the node recommendation, we sample $10\%$ of the nodes from test dataset and evaluate $Precision@k$ and $Recall@k$ for different k values. 
	
	\begin{equation*}
	\displaystyle
	Precision @ k = \frac{| PredSet \cap TestSet |}{| PredSet |} 
	\end{equation*}
	
	\begin{equation*}
	\displaystyle
	Recall @ k = \frac{| PredSet \cap TestSet |}{| TestSet |} 
	\end{equation*}
	
	\hfill \\
	\noindent We report the results of baseline methods and our method on Cora and Epinion datasets in Table \ref{tab:nrcoraresults} and Table \ref{tab:nrepinionresults} respectively. Twitter dataset has about $99\%$ of nodes with zero out-degree, hence, we don't consider it for this task. We can see that, our method improves the precision and recall over DeepWalk for $k = 10$  and $20$ on Cora dataset. On epinion dataset, DeepWalk outperforms all other methods as the dataset has reciprocal edges. Directed graph method, HOPE, NERD, doesn't do well with the reciprocal edges on the recommendation task. GREED does signifacntly well when compared to other directed graph methods in terms of understanding the directed and undirected edges in the graph.
	\begin{table}
		\centering
		\footnotesize
		\begin{tabular}{|p{1.3cm}|*{6}{p{0.65cm}|}}
			\hline
			\multirowcell{3}{Method} & \multicolumn{6}{c|}{Cora} \\
			\cline{2-7}
			& \makecell{P@10} & \makecell{R@10 } & \makecell{P@20 } & \makecell{R@20 } & \makecell{P@50 } & \makecell{R@50 } \\ \hline \hline
			DeepWalk & 0.0472 & 0.2960 & 0.0405 & 0.4734 & 0.0245 & \textbf{0.6730} \\ \hline
			Node2Vec & 0.0290 & 0.1871 & 0.0220 & 0.2656 & 0.0144 & 0.4132 \\ \hline
			LINE & 0.0353 & 0.1984 & 0.0252 & 0.2772 & 0.0150 & 0.3907 \\ \hline
			HOPE & 0.0426 & 0.2193 & 0.0320 & 0.3147 & 0.0189 & 0.4632 \\ \hline
			NERD & 0.0467 & 0.2576 & 0.0350 & 0.3657 & 0.0188 & 0.4665 \\ \hline
			GREED & \textbf{0.0561} & \textbf{0.3408} & \textbf{0.0441} & \textbf{0.4946} & \textbf{0.0254} & 0.6539 \\ \hline
		\end{tabular}
		\caption{Precision and Recall for top-k Node Recommendation on Cora dataset}
		\label{tab:nrcoraresults}
	\end{table}

	\begin{table}
		\centering
		\footnotesize
		\begin{tabular}{|p{1.3cm}|*{6}{p{0.65cm}|}}
			\hline
			\multirowcell{3}{Method} & \multicolumn{6}{c|}{Epinion} \\
			\cline{2-7}
			& \makecell{P@10} & \makecell{R@10 } & \makecell{P@20 } & \makecell{R@20 } & \makecell{P@50 } & \makecell{R@50 } \\ \hline \hline
			DeepWalk & \textbf{0.0627} & \textbf{0.2927} & \textbf{0.0437} & \textbf{0.3486} & \textbf{0.0261} & \textbf{0.4113} \\ \hline
			Node2Vec & 0.0543 & 0.2574 & 0.0387 & 0.3117 & 0.0221 & 0.3685 \\ \hline
			LINE & 0.0201 & 0.0807 & 0.0140 & 0.0961 & 0.0076 & 0.1173 \\ \hline
			HOPE & 0.0144 & 0.0387 & 0.0161 & 0.0682 & 0.0142 & 0.1257 \\ \hline
			NERD & 0.0106 & 0.0261 & 0.0096 & 0.0418 & 0.0079 & 0.0733 \\ \hline
			GREED & 0.0531 & 0.2279 &  0.0392 & 0.2775 & 0.0240 & 0.3335 \\ \hline
		\end{tabular}
		\caption{Precision and Recall for top-k Node Recommendation on Epinion dataset}
		\label{tab:nrepinionresults}
	\end{table}
	
	\section{Conclusion \& Future Work}
	In this paper, we proposed a novel scalable embedding that preserves the edge direction in directed graphs. The algorithm incorporates cross product operation in the realm of deep learning to align embeddings in hyperspace against a reference vector to infer the edge direction. Our experiments support that this embedding in conjunction with a proximity based embedding outperforms the state of the art benchmarks by other methods. \\
	In addition, we believe that there are several extensions to the GREED model that will enhance the applications of embeddings in general and we plan to experiment with the following in the immediate future:     
	\begin{description}
		\item[$\bullet$] Multi-Objective optimization using an improvised neural network architecture to optimize for both proximity and direction to produce a unified embedding for graph nodes
		\item[$\bullet$] Introduce multiple direction vectors $v_{d_1}, v_{d_2},...v_{d_l} $ in the architecture to encode multiple attributes about the node relationships or to produce multi-class predictions about a pair of nodes
		\item[$\bullet$]  Apply GREED to word embeddings to encode additional useful attributes such as antonym/synonym relations among word pairs to help solve NLP use cases better
	\end{description}
	
	\bibliography{references}

\begin{thebibliography}{17}
\providecommand{\natexlab}[1]{#1}
\providecommand{\url}[1]{\texttt{#1}}
\providecommand{\urlprefix}{URL }
\expandafter\ifx\csname urlstyle\endcsname\relax
  \providecommand{\doi}[1]{doi:\discretionary{}{}{}#1}\else
  \providecommand{\doi}{doi:\discretionary{}{}{}\begingroup
  \urlstyle{rm}\Url}\fi

\bibitem[{Bromley et~al.(1994)Bromley, Guyon, LeCun, S{\"a}ckinger, and
  Shah}]{bromley1994signature}
Bromley, J.; Guyon, I.; LeCun, Y.; S{\"a}ckinger, E.; and Shah, R. 1994.
\newblock Signature verification using a" siamese" time delay neural network.
\newblock In \emph{Advances in neural information processing systems},
  737--744.

\bibitem[{Chen et~al.(2007)Chen, Yang, Tang et~al.}]{chen2007directed}
Chen, M.; Yang, Q.; Tang, X.; et~al. 2007.
\newblock Directed Graph Embedding.
\newblock In \emph{IJCAI}, 2707--2712.

\bibitem[{Grover and Leskovec(2016)}]{grover2016node2vec}
Grover, A.; and Leskovec, J. 2016.
\newblock node2vec: Scalable feature learning for networks.
\newblock In \emph{Proceedings of the 22nd ACM SIGKDD international conference
  on Knowledge discovery and data mining}, 855--864.

\bibitem[{Hadsell, Chopra, and LeCun(2006)}]{hadsell2006dimensionality}
Hadsell, R.; Chopra, S.; and LeCun, Y. 2006.
\newblock Dimensionality reduction by learning an invariant mapping.
\newblock In \emph{2006 IEEE Computer Society Conference on Computer Vision and
  Pattern Recognition (CVPR'06)}, volume~2, 1735--1742. IEEE.

\bibitem[{Holland and Leinhardt(1981)}]{holland1981exponential}
Holland, P.~W.; and Leinhardt, S. 1981.
\newblock An exponential family of probability distributions for directed
  graphs.
\newblock \emph{Journal of the american Statistical association} 76(373):
  33--50.

\bibitem[{Khosla et~al.(2019)Khosla, Leonhardt, Nejdl, and
  Anand}]{khosla2019node}
Khosla, M.; Leonhardt, J.; Nejdl, W.; and Anand, A. 2019.
\newblock Node representation learning for directed graphs.
\newblock In \emph{Joint European Conference on Machine Learning and Knowledge
  Discovery in Databases}, 395--411. Springer.

\bibitem[{Kleinberg(1999)}]{kleinberg1999authoritative}
Kleinberg, J.~M. 1999.
\newblock Authoritative sources in a hyperlinked environment.
\newblock \emph{Journal of the ACM (JACM)} 46(5): 604--632.

\bibitem[{Lempel and Moran(2001)}]{lempel2001salsa}
Lempel, R.; and Moran, S. 2001.
\newblock SALSA: the stochastic approach for link-structure analysis.
\newblock \emph{ACM Transactions on Information Systems (TOIS)} 19(2):
  131--160.

\bibitem[{Liben-Nowell and Kleinberg(2007)}]{liben2007link}
Liben-Nowell, D.; and Kleinberg, J. 2007.
\newblock The link-prediction problem for social networks.
\newblock \emph{Journal of the American society for information science and
  technology} 58(7): 1019--1031.

\bibitem[{Mousazadeh and Cohen(2015)}]{mousazadeh2015embedding}
Mousazadeh, S.; and Cohen, I. 2015.
\newblock Embedding and function extension on directed graph.
\newblock \emph{Signal Processing} 111: 137--149.

\bibitem[{Ou et~al.(2016)Ou, Cui, Pei, Zhang, and Zhu}]{ou2016asymmetric}
Ou, M.; Cui, P.; Pei, J.; Zhang, Z.; and Zhu, W. 2016.
\newblock Asymmetric transitivity preserving graph embedding.
\newblock In \emph{Proceedings of the 22nd ACM SIGKDD international conference
  on Knowledge discovery and data mining}, 1105--1114.

\bibitem[{Perozzi, Al-Rfou, and Skiena(2014)}]{perozzi2014deepwalk}
Perozzi, B.; Al-Rfou, R.; and Skiena, S. 2014.
\newblock Deepwalk: Online learning of social representations.
\newblock In \emph{Proceedings of the 20th ACM SIGKDD international conference
  on Knowledge discovery and data mining}, 701--710.

\bibitem[{Perrault-Joncas and Meila(2011)}]{perrault2011directed}
Perrault-Joncas, D.~C.; and Meila, M. 2011.
\newblock Directed graph embedding: an algorithm based on continuous limits of
  laplacian-type operators.
\newblock In \emph{Advances in Neural Information Processing Systems},
  990--998.

\bibitem[{Tang et~al.(2015)Tang, Qu, Wang, Zhang, Yan, and Mei}]{tang2015line}
Tang, J.; Qu, M.; Wang, M.; Zhang, M.; Yan, J.; and Mei, Q. 2015.
\newblock Line: Large-scale information network embedding.
\newblock In \emph{Proceedings of the 24th international conference on world
  wide web}, 1067--1077.

\bibitem[{Wang and Wong(1987)}]{wang1987stochastic}
Wang, Y.~J.; and Wong, G.~Y. 1987.
\newblock Stochastic blockmodels for directed graphs.
\newblock \emph{Journal of the American Statistical Association} 82(397):
  8--19.

\bibitem[{Ying et~al.(2018)Ying, He, Chen, Eksombatchai, Hamilton, and
  Leskovec}]{ying2018graph}
Ying, R.; He, R.; Chen, K.; Eksombatchai, P.; Hamilton, W.~L.; and Leskovec, J.
  2018.
\newblock Graph convolutional neural networks for web-scale recommender
  systems.
\newblock In \emph{Proceedings of the 24th ACM SIGKDD International Conference
  on Knowledge Discovery \& Data Mining}, 974--983.

\bibitem[{Zhou et~al.(2017)Zhou, Liu, Liu, Liu, and Gao}]{zhou2017scalable}
Zhou, C.; Liu, Y.; Liu, X.; Liu, Z.; and Gao, J. 2017.
\newblock Scalable graph embedding for asymmetric proximity.
\newblock In \emph{Proceedings of the Thirty-First AAAI Conference on
  Artificial Intelligence}, 2942--2948.

\end{thebibliography}
	
\end{document}